\documentclass[conference]{IEEEtran}
\IEEEoverridecommandlockouts
\usepackage{cite}
\usepackage{amsmath,amssymb,amsfonts}
\usepackage{algorithmic}
\usepackage{graphicx}
\usepackage{textcomp}
\usepackage{xcolor}
\usepackage{booktabs}
\usepackage{multirow}
\usepackage{multicol}
\usepackage{hyperref}
\usepackage{url}
\def\BibTeX{{\rm B\kern-.05em{\sc i\kern-.025em b}\kern-.08em
    T\kern-.1667em\lower.7ex\hbox{E}\kern-.125emX}}

\begin{document}

\title{Fine-Tuning Large Language Models for Scientific Text Classification: A Comparative Study\\
}

\author{\IEEEauthorblockN{Zhyar Rzgar K Rostam}
\IEEEauthorblockA{\textit{Doctoral School of Applied Informatics and Applied Mathematics} \\
\textit{Óbuda University}\\
Budapest, Hungary \\
zhyar.rostam@stud.uni-obuda.hu}
\and
\IEEEauthorblockN{Gábor Kertész}
\IEEEauthorblockA{\textit{John von Neumann Faculty of Informatics} \\
\textit{Óbuda University}\\
Budapest, Hungary \\
kertesz.gabor@nik.uni-obuda.hu}

}

\maketitle

\begin{abstract}

The exponential growth of online textual content across diverse domains has necessitated advanced methods for automated text classification. Large Language Models (LLMs) based on transformer architectures have shown significant success in this area, particularly in natural language processing (NLP) tasks. However, general-purpose LLMs often struggle with domain-specific content, such as scientific texts, due to unique challenges like specialized vocabulary and imbalanced data. In this study, we fine-tune four state-of-the-art LLMs BERT, SciBERT, BioBERT, and BlueBERT on three datasets derived from the WoS-46985 dataset to evaluate their performance in scientific text classification. Our experiments reveal that domain-specific models, particularly SciBERT, consistently outperform general-purpose models in both abstract-based and keyword-based classification tasks. Additionally, we compare our achieved results with those reported in the literature for deep learning models, further highlighting the advantages of LLMs, especially when utilized in specific domains. The findings emphasize the importance of domain-specific adaptations for LLMs to enhance their effectiveness in specialized text classification tasks. 
\end{abstract}


\section{Introduction}

The digital era has led to an exponential increase in the amount of textual content being shared online daily. This content encompasses a wide array of domains, including scientific literature, political documents, social media posts, and blogs \cite{luo2023exploring, app13063594, amjad2022survey, ahanger2022novel, naseem2022benchmarking, jiao2023brief}. The rapid growth in the volume of this data necessitates the use of Natural Language Processing (NLP) to automate and classify textual information efficiently \cite{naseem2022benchmarking, jiao2023brief, fields2024survey}. Deep learning (DL), as a cutting-edge approach, has demonstrated significant success in this domain \cite{peng2021survey, jiao2023brief, sun2023text}.

Among the various DL architectures, models that utilized transformer architecture achieved better results in recent years. These models have been recognized for their exceptional performance across numerous fields \cite{zhao2023survey, peng2021survey}. Text classification is a fundamental task in NLP, it can be utilized in many applications such as sentiment analysis \cite{alimova2021cross, araci2019finbert, laki2023sentiment}, topic modeling \cite{kherwa2019topic, lezama2023integrating}, information retrieval, and natural language inference. Large Language Models (LLMs), which are built on transformer architectures \cite{vaswani2017attention}, have achieved remarkable success in a wide range of NLP tasks, including text classification \cite{devlin2018bert, zhao2023survey, jiao2023brief, chang2024language, fields2024survey, sun2023text, chae2023large}.

Despite their success, LLMs often face challenges when fine-tuned for specific domains. Scientific texts, in particular, present difficulties due to their specialized vocabulary, distinct grammatical structures, and imbalanced data distributions \cite{beltagy2019scibert, lee2020biobert, peng2019transfer, devlin2018bert, liu2021finbert, dunn2022structured, gupta2022matscibert}. This can result in poor performance when general-purpose LLMs are applied to scientific text classification \cite{devlin2018bert}. The literature highlights the difficulties, emphasizing the need for domain-specific adaptations of LLMs to enhance their effectiveness in specialized areas \cite{beltagy2019scibert, lee2020biobert, peng2019transfer, liu2021finbert, kim2023medibiodeberta}.

To address this issue, we fine-tune four state-of-the-art (SOTA) LLMs ($BERT_{base}$ \cite{devlin2018bert}, $SciBERT_{scivocab}$ \cite{beltagy2019scibert}, $BioBERT_{base}$ \cite{lee2020biobert}, and $BlueBERT_{large}$ \cite{peng2019transfer}) on the WoS-46985 dataset, which consists of 46,985 scientific documents prepared by Kowsari et al. \cite{kowsari2017HDLTex}. We perform two sets of experiments for each model: one using abstracts and another using keywords. In this study, we investigate both general purpose (BERT) and the specific purpose (SciBERT, BioBERT, and BlueBERT) LLMs\footnote{Derived datasets and implementations are available at: \url{https://github.com/ZhyarUoS/Scientific-Text-Classification.git}}. 

The contributions of this study are:
\begin{itemize}
    \item Provide a comprehensive evaluation of domain-specific LLMs (SciBERT, BioBERT, and BlueBERT) in comparison to a general-purpose LLM (BERT), offering valuable benchmarks for future research.
    \item Conduct a systematic evaluation of the impact of using abstracts and keywords as input for LLMs in this context.
    \item Offer a detailed analysis using the WoS-46985 dataset, providing a case study on how domain-specific models can be effectively fine-tuned for scientific text classification.
    \item Provide empirical evidence supporting the superiority of SciBERT for scientific text classification tasks.
    \item Present a comprehensive comparison of our achieved results with those reported in the literature for deep learning models.
\end{itemize}

\section{Related Works}

\subsection{LLMs for Scientific Text Classification}

Beltagy et al. \cite{beltagy2019scibert} present a pre-trained language model (PLM) specifically designed for scientific text. It addresses the challenge of limited high-quality labeled data in the scientific domain by leveraging a massive corpus of scientific publications for unsupervised training. The model significantly outperforms BERT, on various scientific NLP tasks, including sequence tagging, sentence classification, and dependency parsing. This improvement is attributed to SciBERT's specialized training on scientific text. SciBERT is a valuable tool for researchers working with scientific text, offering superior performance compared to general-purpose language models (LM).

Lee et al. \cite{lee2020biobert}  propose a PLM specifically designed for the biomedical domain (BioBERT). The model is built upon the architecture of BERT but is trained on a massive dataset of biomedical text, such as PubMed abstracts and full-text articles. This specialized training allows BioBERT to outperform general-purpose LMs on a variety of biomedical text mining tasks, including named entity recognition (NER) \cite{gou2023lightweight, usha2022named}, relation extraction (RE) \cite{zhao2023comprehensive}, and question answering (QA). BioBERT significantly surpasses previous models in biomedical text mining tasks. This exceptional performance is attributed to its deep understanding of complex medical language and terminology.

SciDeBERTa \cite{beltagy2019scibert} is a PLM specifically tailored for scientific and technological text. The model is built upon the foundation of a general-purpose LM, DeBERTa, and is further refined using a massive dataset of scientific text. This specialized training enables SciDeBERTa to outperform existing models designed for the same purpose, such as SciBERT \cite{beltagy2019scibert} and S2ORC-SciBERT \cite{lo2019s2orc}. The research demonstrates that SciDeBERTa, particularly when fine-tuned for specific domains like computer science (SciDeBERTa-CS), achieves superior performance on tasks such as NER and RE. SciDeBERTa represents a significant advancement in NLP for the scientific and technological domains.

\subsection{Other Deep Learning Approaches for Scientific Text Classification}
HDLTex \cite{kowsari2017HDLTex} provides a hierarchical DL approach for text classification. The model is designed to address the challenges of increasing volume and complexity of document collections. By utilizing a hierarchical structure, HDLTex can effectively classify documents into multiple levels of categories. The model combines different deep learning architectures, such as Deep Neural Networks (DNNs) \cite{schroder2020survey}, Convolutional Neural Networks (CNNs) \cite{lai2015recurrent}, and Recurrent Neural Networks (RNNs) \cite{lee2016sequential, lai2015recurrent}, to capture intricate patterns and relationships within the text data.

\section{Dataset}
\label{dataset}
The dataset utilized for this study is derived from a dataset collected by Kowsari et al. \cite{kowsari2017HDLTex} from the Web of Science (WoS) database and consists of three distinct subsets: WoS-46985, WoS-11967, and WoS-5736 (presented in Tables \ref{tab:46985_abstracts}, \ref{tab:11967_abstracts} and \ref{tab:wos_5736}, respectively). Each dataset varies in size and categorization. The WoS-5736 dataset contains 5,736 documents organized into 11 categories, which are further grouped into 3 parent categories (electrical engineering, psychology, and biochemistry). The WoS-11967 dataset includes 11,967 documents, categorized into 35 categories and grouped under 7 parent categories (computer science, civil engineering, electrical engineering, mechanical engineering, medical sciences, psychology, and biochemistry). The largest of the datasets, WoS-46985, consists of 46,985 documents, divided into 134 categories within the same 7 parent categories.

\begin{table}[tbp]
\caption{WoS-46985: Number of Studies Documents in Different Domains}
    \centering
    \begin{tabular}{ l  c }
        \hline
        \textbf{Domain} & \textbf{Number of Abstracts} \\ \hline
        Computer Science & 6514 \\ \hline
        Civil Engineering & 4237 \\ \hline
        Electrical Engineering & 5483 \\ \hline
        Mechanical Engineering & 3297 \\ \hline
        Medical Sciences & 14625 \\ \hline
        Psychology & 7142 \\ \hline
        Biochemistry & 5687 \\ \hline
        \textbf{Total} & \textbf{46985} \\ \hline
    \end{tabular}
    \label{tab:46985_abstracts}
\end{table}

\begin{table}[tbp]
 \caption{WoS-11967: Number of Documents in Different Domains}
    \centering
    \begin{tabular}{ l  c }
        \hline
        \textbf{Domain} & \textbf{Number of Abstracts} \\ \hline
        Computer Science & 1499 \\ \hline
        Civil Engineering & 2107 \\ \hline
        Electrical Engineering & 1132 \\ \hline
        Mechanical Engineering & 1925 \\ \hline
        Medical Sciences & 1617 \\ \hline
        Psychology & 1959 \\ \hline
        Biochemistry & 1728 \\ \hline
        \textbf{Total} & \textbf{11967} \\ \hline
    \end{tabular}
    \label{tab:11967_abstracts}
\end{table}

\begin{table}[tbp]
    \caption{WoS-5736: Number of Documents in Different Domains}
    \centering
    \begin{tabular}{ l  c }
        \hline
        \textbf{Domain} & \textbf{Number of Abstracts} \\ \hline
        Electrical Engineering & 1292 \\ \hline
        Psychology & 1597 \\ \hline
        Biochemistry & 2847 \\ \hline
        \textbf{Total} & \textbf{5736} \\ \hline
    \end{tabular}
    \label{tab:wos_5736}
\end{table}

\section{Methods}
\subsection{Dataset Preparation and Preprocessing}
\label{data_preperation}
Each dataset (WoS-5736, WoS-11967, and WoS-46985) underwent a structured preparation process to extract four primary attributes: Labels, Domains, Keywords, and Abstracts. Metadata from the original WoS datasets was meticulously examined to identify common studies, from which the desired fields were extracted. Subsequently, the following preprocessing steps were applied to the extracted data and stored in a Tab-Separated Values (TSV) format:
\begin{itemize}
    \item Removal of extra spaces: Unnecessary spaces within domain labels were eliminated.
    \item Textual data was converted to lowercase and stripped of non-alphanumeric characters (except spaces).    
\end{itemize}
Furthermore, the dataset randomized to mitigate potential biases. Subsequently, we partitioned the datasets into training (80\%), testing (20\%), and validation (20\% of the test set) subsets. To ensure consistency in data handling, all experiments adhered to this standardized data split, and presented in Table \ref{tab:dataset_splits}.

\begin{table}[tbp]
    \centering
    \caption{Dataset Splits for WoS Datasets}
    \label{tab:dataset_splits}
    \begin{tabular}{lrrr}
        \toprule
        \textbf{Dataset} & \textbf{Train} & \textbf{Test} & \textbf{Validation} \\
        \midrule
        WoS-5736  & 4588  & 1148  & 230  \\
        WoS-11967 & 9573  & 2394  & 479  \\
        WoS-46985 & 37588 & 9397  & 1880 \\
        \bottomrule
    \end{tabular}
\end{table}
\subsection{Data Tokenization and Encoding}
\label{data_tkenization}
To facilitate model training, the textual data (abstracts, and keywords) were transformed into numerical representations. This process involved tokenization, where text is broken down into smaller units (tokens), and encoding, where tokens are mapped to numerical values. We utilized a tokenizer with respect to the models. The tokenizer converted text sequences into input IDs and attention masks, essential for model input.

\subsection{Experimental Design}
\label{expermintal_design}
To comprehensively evaluate the performance of various LMs, two experimental setups were implemented for each model. In the first experiment, the model was trained and evaluated using only the abstract of each scientific document. In the second experiment we focused on utilizing only the keywords associated with the document. This comparative approach allowed for a thorough assessment of the models' capabilities in handling different textual representations.

A range of PLMs, including both general-purpose (BERT) and domain-specific (SciBERT, BioBERT, and BlueBERT) models, were included in the study. This diverse model selection enabled a comparative analysis of their performance in scientific text classification. By investigating the impact of different text representations (abstracts vs. keywords) and model architectures, this study aimed to identify the most effective approach for this specific task.

To ensure a fair comparison across all models, a standardized fine-tuning process was adopted and executed on Google Colab using a T4 GPU. The AdamW optimizer was employed with a learning rate of \(2 \times 10^{-5}\) and epsilon of \(1 \times 10^{-8}\). A linear learning rate scheduler with warmup was utilized, commencing with a warmup period of \(1 \times 10^{-4}\) steps. The models underwent training for a total of 20 epochs (a summary presented in Table \ref{tab:training_parameters}). These consistent training parameters facilitated a focused evaluation of the models' performance based on their underlying architectures and the nature of the input data (abstracts or keywords).

\begin{table}[tbp]
    \centering
    \caption{Training Configuration Parameters}
    \label{tab:training_parameters}
    \begin{tabular}{lr}
        \toprule
        \textbf{Parameter} & \textbf{Value} \\
        \midrule
        Optimizer          & AdamW \\
        Learning Rate      & \(2 \times 10^{-5}\) \\
        Epsilon            & \(1 \times 10^{-8}\) \\
        Scheduler          & Linear with warmup \\
        Warmup Steps       & \(1 \times 10^{-4}\) \\
        Epochs & \( \text{20} \) \\
        \bottomrule
    \end{tabular}
\end{table}

\section{Results}
This section presents the model's performance and efficiency on each scenario individually and then reports the best achieved results among experimented LLMs. All models' performance evaluations are presented in Table \ref{tab:model_perfomance_metrics}.

\subsection{WoS-46985: Abstracts}
Among the models evaluated, SciBERT demonstrated the highest performance on the WoS-46985 dataset, achieving an accuracy of 87\% and consistently higher F1 scores compared to BERT, BioBERT, and BlueBERT. While BlueBERT and BioBERT both achieved an accuracy of 86\%, SciBERT's superior precision, recall, and F1 balance across classes suggest its suitability for scientific text classification. BioBERT and BlueBERT, which are tailored for biomedical contexts, displayed comparable performance to BERT, with slight variability in F1 scores, but did not surpass SciBERT (see Fig. \ref{fig:46985_both}).

\subsection{WoS-46985: Keywords}
\begin{figure}[tbp]
\centerline{\includegraphics[width=.5\textwidth]{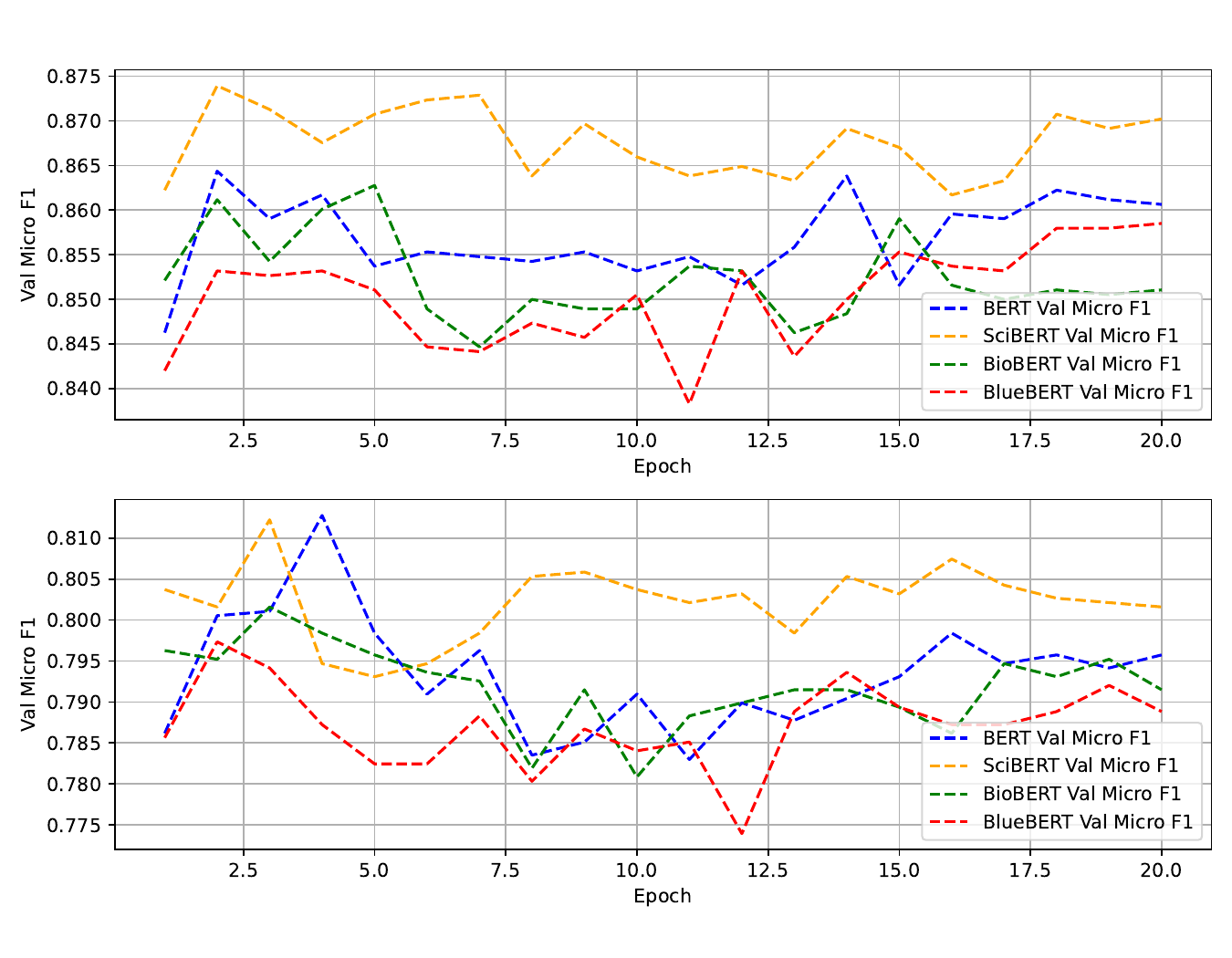}}
\caption{Performance evaluation on the WoS-46985 dataset for BERT, SciBERT, BioBERT, and BlueBERT LMs. The top sub-figure shows the evolution of models when utilizing \textit{abstracts}, while the bottom sub-figure shows the evolution of models when utilizing \textit{keywords}, as measured by the F1 score on the validation subset.}
\label{fig:46985_both}
\end{figure}

SciBERT and BlueBERT consistently outperformed the other models while fine-tuning with WoS-46985 dataset and utilizing keywords as input. The classification reports reveal that SciBERT and BlueBERT also delivered superior precision, recall, and F1-scores across most categories. The results highlight both BlueBERT and SciBERT performance in classification tasks (see Fig. \ref{fig:46985_both}), particularly in the biomedical domain, with BioBERT and BERT following closely. 

\subsection{WoS-11967: Abstracts}
Our experiments show that BERT achieved a notable peak validation Micro F1 score of 0.92 by the 20\textsuperscript{th} epoch, with a final classification accuracy of 91\% while we use abstracts from WoS-11967 as an input. SciBERT reached a maximum accuracy of 92\%, demonstrating slightly better performance in classification tasks. While all models showed high performance, SciBERT slightly outperformed the others in terms of F1 score and accuracy, emphasizing their potential advantages in specific domains of text classification (details presented in Fig. \ref{fig:11967_both}).

\subsection{WoS-11967: Keywords}
\begin{figure}[tbp]
\centerline{\includegraphics[width=.5\textwidth]{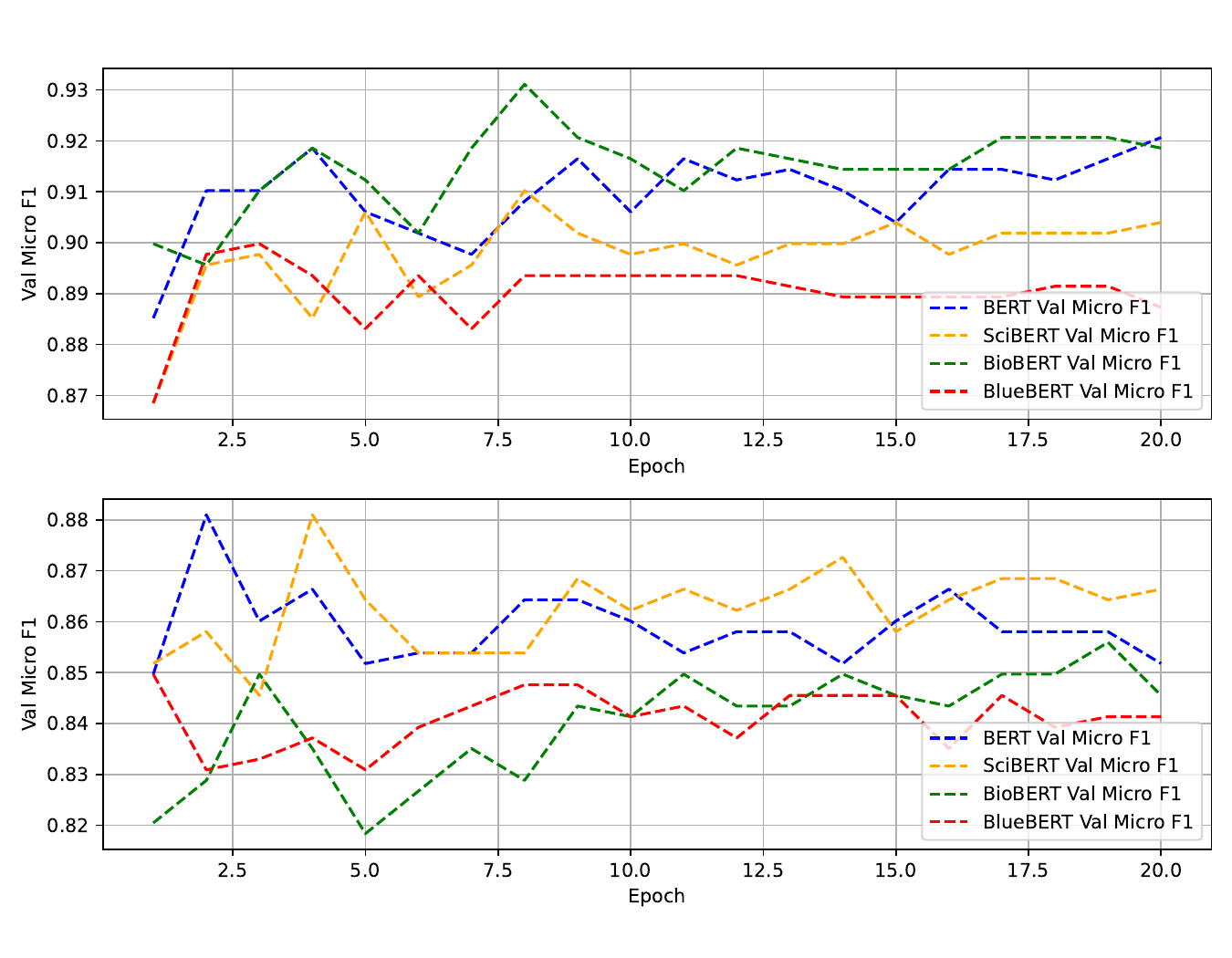}}
\caption{Performance evaluation on the WoS-11967 dataset for BERT, SciBERT, BioBERT, and BlueBERT LMs. The top sub-figure shows the evolution of models when utilizing \textit{abstracts}, while the bottom sub-figure shows the evolution of models when utilizing \textit{keywords}, as measured by the F1 score on the validation subset.}
\label{fig:11967_both}
\end{figure}

In the case of fine-tuning models with WoS-11967 (keywords), BERT achieved a final validation micro F1 score of 0.85 with a test accuracy of 84\%. SciBERT demonstrated superior performance with a final micro F1 score of 0.87 and a test accuracy of 87\%. In comparison, BioBERT reached a final micro F1 score of 0.85 and an accuracy of 86\%. As a result, SciBERT outperformed the other models in both F1 score and accuracy, indicating its better effectiveness for the given classification task (model's performance presented in Fig. \ref{fig:11967_both}).

\subsection{WoS-5736: Abstracts}
In the experiments WoS-5736 dataset (abstract as input) with BERT, SciBERT, BioBERT, and BlueBERT, all models achieved high performance in text classification tasks. BERT demonstrated steady improvements in validation micro F1 scores, reaching 0.98 by the final epoch, with a final accuracy of 97\%. SciBERT also showed consistent enhancement in validation micro F1 scores, peaking at 0.97, and achieved an overall accuracy of 98\%. BioBERT exhibited high performance with a final micro F1 score of 0.99 and an impressive accuracy of 98\%. BlueBERT, despite its longer training time, achieved a good validation micro F1 score of 0.92 and an overall accuracy of 96\%. To ensure a fair comparison, all models were trained for 20 epochs. However, it is important to note that each model achieved its peak performance prior to the 10\textsuperscript{th} epoch. (see Fig. \ref{fig:5736_both}).

\subsection{WoS-5736: Keywords}
\begin{figure}[tbp]
\centerline{\includegraphics[width=.5\textwidth]{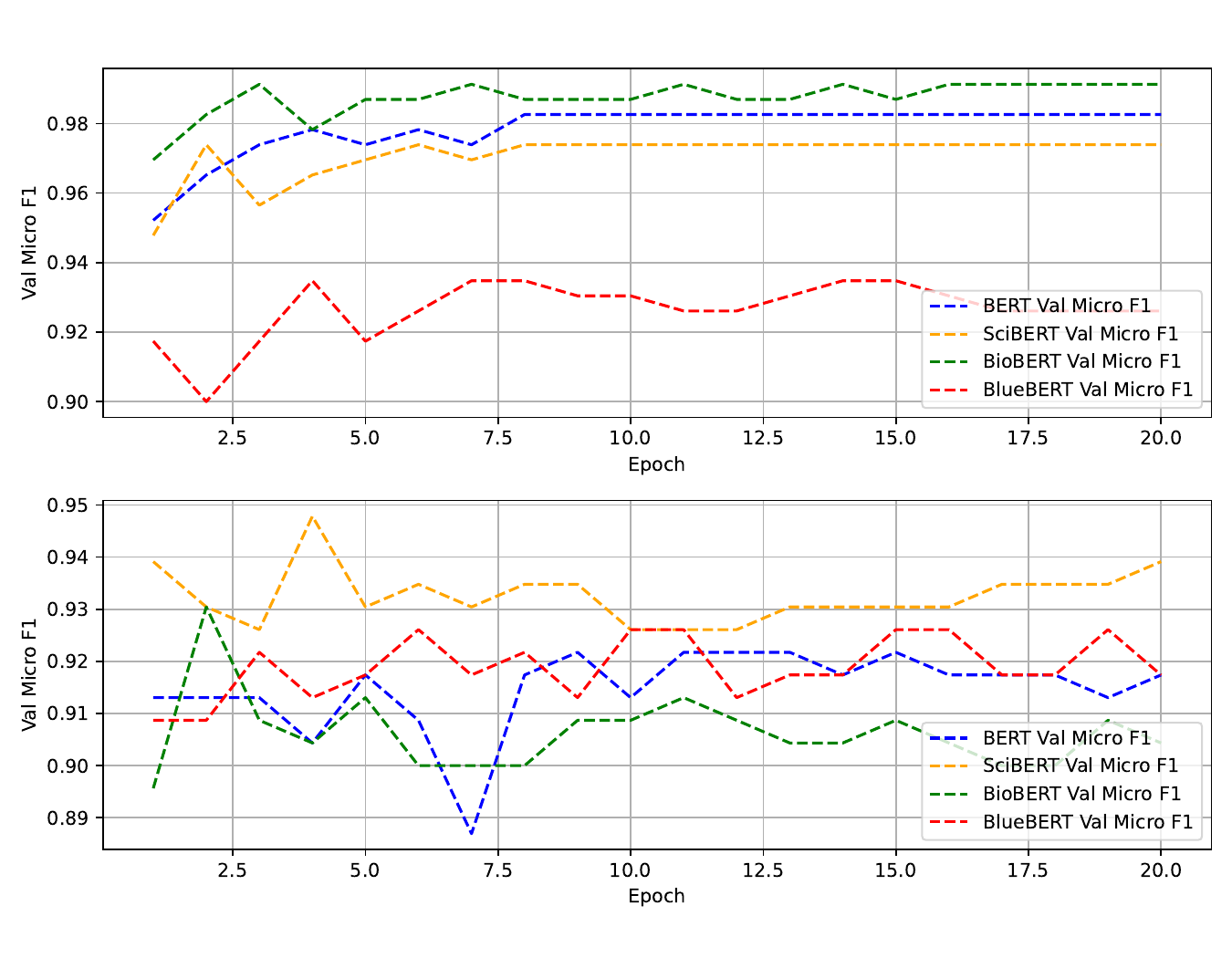}}
\caption{Performance evaluation on the WoS-5736 dataset for BERT, SciBERT, BioBERT, and BlueBERT LMs. The top sub-figure shows the evolution of models when utilizing \textit{abstracts}, while the bottom sub-figure shows the evolution of models when utilizing \textit{keywords}, as measured by the F1 score on the validation subset.}
\label{fig:5736_both}
\end{figure}

In our final experiment, BERT achieved a peak validation Micro F1 score of 0.92 with a final accuracy of 93\%, while SciBERT reached a maximum Micro F1 score of 0.94 and an accuracy of 94\%. BioBERT's highest Micro F1 score was 0.93 with a final accuracy of 93\%, and BlueBERT attained an accuracy of 93\%. SciBERT generally performed best, achieving the highest validation scores consistently, while other models showed competitive results (model's performance evaluation presented in Fig. \ref{fig:5736_both}).

\begin{table}[tbp]
\centering
\caption{Models Performance Evaluation}
\label{tab:model_perfomance_metrics}
\scalebox{0.6}
{
\begin{tabular}{l|ll|ll|ll|l}
\toprule

\textbf{Models} & \textbf{F Scores} & & \textbf{Recall Scores} & & \textbf{Precision Scores} & & \textbf{Accuracy}
\\
\midrule
\multicolumn{7}{c}{WoS-46985: Abstracts}
\\
\midrule

\multirow{3}{*}{\textbf{BERT}} & & & & & & & \multirow{3}{*}{85\%}  \\
 & Macro F1 & 0.8496 & Macro Recall & 0.8501 & Macro Precision & 0.8494 & \\
 & Micro F1 & 0.8542 & Micro Recall & 0.8542 & Micro Precision & 0.8542 & \\
 & Weighted F1 & 0.8541 & Weighted Recall & 0.8542 & Weighted Precision & 0.8543 & \\
\midrule

\multirow{3}{*}{\textbf{SciBERT}} & & & & & & & \multirow{3}{*}{\textbf{87\%}} \\
  & Macro F1 & \textbf{0.8666} & Macro Recall & \textbf{0.8657} & Macro Precision & \textbf{0.8676} & \\
 & Micro F1 & \textbf{0.8691} & Micro Recall & \textbf{0.8691}& Micro Precision & \textbf{0.8691} & \\
 & Weighted F1 & \textbf{0.8688} & Weighted Recall & \textbf{0.8691} & Weighted Precision & \textbf{0.8688} & \\
 
\midrule
\multirow{3}{*}{\textbf{BioBERT}} & & & & & & & \multirow{3}{*}{86\%} \\
 & Macro F1 & 0.8557 & Macro Recall & 0.8541 & Macro Precision & 0.8574 & \\
 & Micro F1 & 0.8566 & Micro Recall & 0.8566 & Micro Precision & 0.8566 & \\
 & Weighted F1 & 0.8568 & Weighted Recall & 0.8566 & Weighted Precision & 0.8571 & \\
\midrule
\multirow{3}{*}{\textbf{BlueBERT}} & & & & & & & \multirow{3}{*}{86\%} \\
 & Macro F1 & 0.8545 & Macro Recall & 0.8528 & Macro Precision & 0.8564 & \\
 & Micro F1 & 0.8566 & Micro Recall & 0.8566 & Micro Precision & 0.8566 & \\
 & Weighted F1 & 0.8566 & Weighted Recall & 0.8566 & Weighted Precision & 0.8568 & \\
\bottomrule

% WoS-46985: Keywords
\midrule
\multicolumn{7}{c}{WoS-46985: Keywords}
\\
\midrule
\multirow{3}{*}{\textbf{BERT}} & & & & & & & \multirow{3}{*}{79\%}  \\
 & Macro F1 & 0.7789 & Macro Recall & 0.7780 & Macro Precision & 0.7807 & \\
 & Micro F1 & 0.7944 & Micro Recall & 0.7944 & Micro Precision & 0.7944 & \\
 & Weighted F1 & 0.7939 & Weighted Recall & 0.7944 & Weighted Precision & 0.7940 & \\
\midrule
\multirow{3}{*}{\textbf{SciBERT}} & & & & & & & \multirow{3}{*}{\textbf{80\%}}  \\
 & Macro F1 & 0.7830 & Macro Recall & \textbf{0.7818} & Macro Precision & 0.7845 & \\
 & Micro F1 & 0.7951 & Micro Recall & 0.7951 & Micro Precision & \textbf{0.7951} & \\
 & Weighted F1 & 0.7950 & Weighted Recall & 0.7951 & Weighted Precision & 0.7952 & \\
\midrule
\multirow{3}{*}{\textbf{BioBERT}} & & & & & & & \multirow{3}{*}{79\%} \\
 & Macro F1 & 0.7836 & Macro Recall & 0.7815 & Macro Precision & 0.7818 & \\
 & Micro F1 & 0.7949 & Micro Recall & 0.7949 & Micro Precision & 0.7949 & \\
 & Weighted F1 & 0.7944 & Weighted Recall & 0.7949 & Weighted Precision & 0.7942 & \\
\midrule
\multirow{3}{*}{\textbf{BlueBERT}} & & & & & & & \multirow{3}{*}{\textbf{80\%}} \\
 & Macro F1 & \textbf{0.7854} & Macro Recall & 0.7814 & Macro Precision & \textbf{0.7879} & \\
 & Micro F1 & \textbf{0.7987} & Micro Recall & \textbf{0.7987} & Micro Precision & 0.7879 & \\
 & Weighted F1 & \textbf{0.7980} & Weighted Recall & \textbf{0.7987} & Weighted Precision & \textbf{0.7979} & \\
\bottomrule

% 11967:Abstracts
\midrule
\multicolumn{7}{c}{WoS-11967: Abstracts}
\\
\midrule
\multirow{3}{*}{\textbf{BERT}} & & & & & & & \multirow{3}{*}{91\%}  \\
 & Macro F1 & 0.9031 & Macro Recall & 0.9044 & Macro Precision & 0.9023 & \\
 & Micro F1 & 0.9060 & Micro Recall & 0.9060 & Micro Precision & 0.9060 & \\
 & Weighted F1 & 0.9060 & Weighted Recall & 0.9060 & Weighted Precision & 0.9065 & \\
\midrule
\multirow{3}{*}{\textbf{SciBERT}} & & & & & & & \multirow{3}{*}{\textbf{92\%}}  \\
 & Macro F1 & \textbf{0.9205} & Macro Recall & \textbf{0.9222} & Macro Precision & \textbf{0.9193} & \\
 & Micro F1 & \textbf{0.9218} & Micro Recall & \textbf{0.9218} & Micro Precision & \textbf{0.9218} & \\
 & Weighted F1 & \textbf{0.9218} & Weighted Recall & \textbf{0.9218} & Weighted Precision & \textbf{0.9222} & \\
\midrule
\multirow{3}{*}{\textbf{BioBERT}} & & & & & & & \multirow{3}{*}{91\%} \\
 & Macro F1 & 0.9034 & Macro Recall & 0.9024 & Macro Precision & 0.9048 & \\
 & Micro F1 & 0.9055 & Micro Recall & 0.9055 & Micro Precision & 0.9055 & \\
 & Weighted F1 & 0.9055 & Weighted Recall & 0.9055 & Weighted Precision & 0.9058 & \\
\midrule
\multirow{3}{*}{\textbf{BlueBERT}} & & & & & & & \multirow{3}{*}{91\%} \\
 & Macro F1 & 0.9060 & Macro Recall & 0.9078 & Macro Precision & 0.9046 & \\
 & Micro F1 & 0.9085 & Micro Recall & 0.9085 & Micro Precision & 0.9085 & \\
 & Weighted F1 & 0.9087 & Weighted Recall & 0.9085 & Weighted Precision & 0.9092 & \\
\bottomrule

% 11967:Keywords
\midrule
\multicolumn{7}{c}{WoS-11967: Keywords}
\\
\midrule
\multirow{3}{*}{\textbf{BERT}} & & & & & & & \multirow{3}{*}{84\%}  \\
 & Macro F1 & 0.8369 & Macro Recall & 0.8365 & Macro Precision & 0.8384 & \\
 & Micro F1 & 0.8421 & Micro Recall & 0.8421 & Micro Precision & 0.8421 & \\
 & Weighted F1 & 0.8418 & Weighted Recall & 0.8421 & Weighted Precision & 0.8423 & \\
\midrule
\multirow{3}{*}{\textbf{SciBERT}} & & & & & & & \multirow{3}{*}{\textbf{87\%}}  \\
 & Macro F1 & \textbf{0.8693} & Macro Recall & \textbf{0.8689} & Macro Precision & \textbf{0.8704} & \\
 & Micro F1 & \textbf{0.8730} & Micro Recall & \textbf{0.8730} & Micro Precision & \textbf{0.8730} & \\
 & Weighted F1 & \textbf{0.8704} & Weighted Recall & \textbf{0.8730} & Weighted Precision & \textbf{0.8733} & \\
\midrule
\multirow{3}{*}{\textbf{BioBERT}} & & & & & & & \multirow{3}{*}{86\%} \\
 & Macro F1 & 0.8518 & Macro Recall & 0.8521 & Macro Precision & 0.8528 & \\
 & Micro F1 & 0.8554 & Micro Recall & 0.8554 & Micro Precision & 0.8554 & \\
 & Weighted F1 & 0.8553 & Weighted Recall & 0.8554 & Weighted Precision & 0.8564 & \\
\midrule
\multirow{3}{*}{\textbf{BlueBERT}} & & & & & & & \multirow{3}{*}{85\%} \\
 & Macro F1 & 0.8486 & Macro Recall & 0.8485 & Macro Precision & 0.8486 & \\
 & Micro F1 & 0.8521 & Micro Recall & 0.8521 & Micro Precision & 0.8521 & \\
 & Weighted F1 & 0.8521 & Weighted Recall & 0.8521 & Weighted Precision & 0.8521 & \\
\bottomrule

% WoS-5736:Abstract

\midrule
\multicolumn{7}{c}{WoS-5736: Abstracts}
\\
\midrule
\multirow{3}{*}{\textbf{BERT}} & & & & & & & \multirow{3}{*}{97\%}  \\
 & Macro F1 & 0.9649 & Macro Recall & 0.9618 & Macro Precision & 0.9687 & \\
 & Micro F1 & 0.9684 & Micro Recall & 0.9684 & Micro Precision & 0.9686 & \\
 & Weighted F1 & 0.9684 & Weighted Recall & 0.9684 & Weighted Precision & 0.9687 & \\
\midrule
\multirow{3}{*}{\textbf{SciBERT}} & & & & & & & \multirow{3}{*}{\textbf{98}\%}  \\
 & Macro F1 & 0.9739 & Macro Recall & 0.9715 & Macro Precision & \textbf{0.9763} & \\
 & Micro F1 & 0.9756 & Micro Recall & 0.9756 & Micro Precision & 0.9756 & \\
 & Weighted F1 & 0.9755 & Weighted Recall & 0.9756 & Weighted Precision & 0.9756 & \\
\midrule
\multirow{3}{*}{\textbf{BioBERT}} & & & & & & & \multirow{3}{*}{\textbf{98\%}} \\
 & Macro F1 & \textbf{0.9749} & Macro Recall & \textbf{0.9747} & Macro Precision & 0.9750 & \\
 & Micro F1 & \textbf{0.9773 }& Micro Recall & \textbf{0.9773} & Micro Precision & \textbf{0.9773} & \\
 & Weighted F1 & \textbf{0.9773} & Weighted Recall & \textbf{0.9773} & Weighted Precision & \textbf{0.9773} & \\
\midrule
\multirow{3}{*}{\textbf{BlueBERT}} & & & & & & & \multirow{3}{*}{96\%} \\
 & Macro F1 & 0.9540 & Macro Recall & 0.9510 & Macro Precision & 0.9572 & \\
 & Micro F1 & 0.9581 & Micro Recall & 0.9581 & Micro Precision & 0.9581 & \\
 & Weighted F1 & 0.9579 & Weighted Recall & 0.9581 & Weighted Precision & 0.9580 & \\
\bottomrule

% 5736:Keywords
\midrule
\multicolumn{7}{c}{WoS-5736: Keywords}
\\
\midrule
\multirow{3}{*}{\textbf{BERT}} & & & & & & & \multirow{3}{*}{93\%}  \\
 & Macro F1 & 0.9248 & Macro Recall & 0.9213 & Macro Precision & 0.929 & \\
 & Micro F1 & 0.9329 & Micro Recall & 0.9329 & Micro Precision & 0.9329 & \\
 & Weighted F1 & 0.9323 & Weighted Recall & 0.9329 & Weighted Precision & 0.9323 & \\
\midrule
\multirow{3}{*}{\textbf{SciBERT}} & & & & & & & \multirow{3}{*}\textbf{{94\%}} \\
 & Macro F1 & \textbf{0.9373} & Macro Recall & \textbf{0.9387} & Macro Precision & \textbf{0.9359} & \\
 & Micro F1 & \textbf{0.9416} & Micro Recall & \textbf{0.9416} & Micro Precision & \textbf{0.9416} & \\
 & Weighted F1 & \textbf{0.9416} & Weighted Recall & \textbf{0.9416} & Weighted Precision & \textbf{0.9417} & \\
\midrule
\multirow{3}{*}{\textbf{BioBERT}} & & & & & & & \multirow{3}{*}{93\%} \\
 & Macro F1 & 0.9165 & Macro Recall & 0.9167 & Macro Precision & 0.9163 & \\
 & Micro F1 & 0.9259 & Micro Recall & 0.9259 & Micro Precision & 0.9259 & \\
 & Weighted F1 & 0.9257 & Weighted Recall & 0.9259 & Weighted Precision & 0.9256 & \\
\midrule
\multirow{3}{*}{\textbf{BlueBERT}}& & & & & & & \multirow{3}{*}{93\%} \\
 & Macro F1 & 0.9223 & Macro Recall & 0.9215 & Macro Precision & 0.9241 & \\
 & Micro F1 & 0.9303 & Micro Recall & 0.9303 & Micro Precision & 0.9303 & \\
 & Weighted F1 & 0.9297 & Weighted Recall & 0.9303 & Weighted Precision & 0.9299 & \\
\bottomrule

\end{tabular}
}
\end{table}

\section{Discussion}
In this section, we provide a discussion with a comparison among the achieved results while utilizing LLMs against results reported in the literature \cite{kowsari2017HDLTex} (details presented in Table \ref{tab:models_accuracies}).

\begin{table}[tbp]
\caption{LLMs Accuracy Against Other Deep Learning Approaches}
\label{tab:models_accuracies}
\small
\setlength{\tabcolsep}{4pt} % reduce column spacing
\renewcommand{\arraystretch}{0.7} % reduce row height
\begin{tabular}{l|l|l|l|l}
\multirow{2}{*}{}            & \multicolumn{2}{c}{WoS-11967}               & WoS-46985 & WoS-5736 \\
\midrule
                             & Methods & Accuracy     & Accuracy  & Accuracy \\
 \midrule                            
Baseline &   DNN           & 80.02             &     66.95          & 86.15  \\
                             &   CNN      &  83.29            & 70.46              & 88.68 \\
                             &   RNN            &  83.96            &   72.12            & 89.46  \\
                             &   NBC          &      68.8        &       46.2        & 78.14         \\
                             &   SVM            &  80.65            &   67.56            & 85.54  \\
                             &   SVM           &    83.16          &    70.22           & 88.24  \\
                             &   Stacking SVM          &   79.45           &     71.81          & 85.68       \\
\midrule                             
HDLTex                        &  HDLTex  &  86.07          &  76.58            &    90.93     \\                           
 \midrule   
 \multirow{2}{*}{}            & \multicolumn{2}{c}{WoS-46985}               & WoS-11967 & WoS-5736 \\
  \midrule  
LLMs:  &   BERT           & 85.0             &     91.0          & 96.0  \\   
Abstracts&   SciBERT        & \textbf{87.0}             &     \textbf{92.0}          & 97.0  \\ 
                 &   BioBERT        & 86.0             &     91.0          & \textbf{98.0}  \\
                 &   BlueBERT       & 86.0             &     91.0          & 97.0  \\ 
\midrule 
LLMs:   &   BERT           & 79.0             &     84.0          & 93.0  \\   
Keywords    &   SciBERT        & \textbf{80.0}             &     \textbf{87.0}          & \textbf{94.0}  \\ 
                 &   BioBERT        & 79.0             &     86.0          & 93.0  \\
                 &   BlueBERT       & \textbf{80.0}             &     85.0          & 93.0  \\ 
\midrule 
\end{tabular}
\end{table}

Based on our results and a comparison with existing literature, our models consistently outperformed baseline models and the HDLTex model when utilizing abstracts. However, achieving high classification performance when feeding the model only keywords is a challenging task. Despite this, our setup with LLMs outperformed the baselines and HDLTex in most cases. Notably, SciBERT demonstrated superior performance in scientific and domain-specific text classification tasks across various WoS datasets, consistently surpassing other models such as BERT, BioBERT, and BlueBERT in terms of accuracy, precision, recall, and F1 scores.

Moreover, on the WoS-46985 dataset, SciBERT achieved the highest accuracy and F1 scores, highlighting its robustness in scientific text classification. When using keywords as input, SciBERT maintained its leading position, delivering the highest validation micro F1 scores across all datasets. While BlueBERT exhibited competitive performance in later epochs, it was less consistent compared to SciBERT. BioBERT and BERT also performed well, particularly in the biomedical domain, but their results did not outperform SciBERT.

These findings suggest that SciBERT's domain-specific optimizations significantly enhance its effectiveness in specialized text classification tasks. Although BioBERT and BlueBERT showed strengths in certain contexts, SciBERT's consistent performance across diverse datasets underscores its potential as the most reliable model for scientific and technical text classification.

\section{Conclusion and Future Directions}
This study demonstrates the critical role of domain-specific adaptations in enhancing the performance of LLMs for scientific text classification. Our experiments highlight SciBERT's consistent superiority over both general-purpose and other domain-specific models, particularly in handling abstracts and keywords across various datasets derived from the WoS-46985 dataset. The results indicate that fine-tuning LLMs on domain-specific corpora significantly improves their ability to manage the complexities of specialized texts, such as those found in scientific literature.

There are several directions for future research. First, exploring further fine-tuning techniques, such as continual learning and domain-adaptive pertaining, could achieve better performance in domain-specific tasks. Additionally, expanding the scope of datasets to include more diverse and larger scientific corpora could test the models' scalability and robustness. Furthermore, investigating the impact of different data preprocessing techniques, and hyperparameter optimization is essential.

\section{Limitations}
While this study highlights the effectiveness of domain-specific LLMs, it has several limitations:
\begin{itemize}
    \item The study is limited to the WoS dataset, which primarily focuses on scientific texts; therefore, the results may not be generalizable to other domains or types of textual data.
    \item Due to limited access to powerful computing resources fine-tuning process was performed using a standardized set of hyperparameters, which may not have been optimal for all models or datasets.
    \item The experiments were conducted using only abstracts and keywords, which may not capture the full complexity of the documents.
\end{itemize}

\section{Acknowledgement}
The authors express their gratitude to the members of the Applied Machine Learning Research Group at Óbuda University's John von Neumann Faculty of Informatics for their valuable comments and suggestions. They also wish to acknowledge the support provided by the Doctoral School of Applied Informatics and Applied Mathematics at Óbuda University.
\bibliographystyle{ieeetr}
\bibliography{ref}

\end{document}